\documentclass[preprint,12pt,authoryear]{elsarticle}

\usepackage{amssymb}
\usepackage{amsmath}
\usepackage{rotating}
\usepackage{algorithm, algorithmic}
\usepackage{multirow}
\usepackage{graphicx}
\usepackage{lineno}

\begin{document}

\makeatletter
\def\ps@pprintTitle{%
   \let\@oddhead\@empty
   \let\@evenhead\@empty
   \let\@oddfoot\@empty
   \let\@evenfoot\@empty
}
\makeatother

\begin{frontmatter}



\title{TransGAT: Transformer-Based Graph Neural Networks for Multi-Dimensional Automated Essay Scoring} %% Article title


\author[1]{Hind Aljuaid\corref{cor1}}
\cortext[cor1]{Corresponding author}
\ead{hhamadaljuaid0001@stu.kau.edu.sa}

\author[1]{Areej Alhothali}
\author[1]{Ohoud Al-Zamzami} %% Author name

\affiliation[1]{organization={Department of Computer Science, King Abdulaziz University}, %Department and Organization
             city={Jeddah}, 
             state={Makkah}, 
             postcode={21589}, 
             country={Saudi Arabia}}

\author[2]{Hussein Assalahi} %% Author name

\affiliation[2]{organization={English Language Institute, King Abdulaziz University}, %Department and Organization
             city={Jeddah}, 
             state={Makkah}, 
             postcode={21589}, 
             country={Saudi Arabia}}

%\author[3]{Tahani Aldosemani} %% Author name

%\affiliation[3]{organization={College of Education, Prince Sattam Bin Abdulaziz University}, %Department and Organization
%             city={Al-Kharj}, 
%             state={Riyadh}, 
%             postcode={16273}, 
%             country={Saudi Arabia}}
%% Abstract
\begin{abstract}
Essay writing is a critical component of student assessment, yet manual scoring is labor-intensive and inconsistent. Automated Essay Scoring (AES) offers a promising alternative, but current approaches face limitations. Recent studies have incorporated Graph Neural Networks (GNNs) into AES using static word embeddings that fail to capture contextual meaning, especially for polysemous words. Additionally, many methods rely on holistic scoring, overlooking specific writing aspects such as grammar, vocabulary, and cohesion. To address these challenges, this study proposes TransGAT, a novel approach that integrates fine-tuned Transformer models with GNNs for analytic scoring. TransGAT combines the contextual understanding of Transformers with the relational modeling strength of Graph Attention Networks (GAT). It performs two-stream predictions by pairing each fine-tuned Transformer (BERT, RoBERTa, and DeBERTaV3) with a separate GAT. In each pair, the first stream generates essay-level predictions, while the second applies GAT to Transformer token embeddings, with edges constructed from syntactic dependencies. The model then fuses predictions from both streams to produce the final analytic score. Experiments on the ELLIPSE dataset show that TransGAT outperforms baseline models, achieving an average Quadratic Weighted Kappa (QWK) of 0.854 across all analytic scoring dimensions. These findings highlight the potential of TransGAT to advance AES systems.
\end{abstract}

% \includegraphics[width=\textwidth]{The analytic method - last.pdf}

% \begin{highlights}
% \item Proposes a hybrid model combining Transformer and Graph Attention Networks for essay scoring.
% \item Evaluates essays on multiple dimensions such as cohesion, syntax, and grammar.
% \item Demonstrates the effectiveness of attention-based graph modeling in AES tasks.
% Achieves improved accuracy over existing AES approaches.
% \item Demonstrates effectiveness on benchmark dataset for second language learners.
% \end{highlights}

%% Keywords

\end{frontmatter}

% \linenumbers


\section{Introduction}
\label{sec:introduction}
Essay writing tasks are commonly used in educational settings to assess a student's creativity, critical thinking, and subject knowledge. Such tasks require the ability to collect, synthesize, and present ideas and arguments \citep{west2019writing}. Essay-based assessments are integral to a variety of contexts, including classroom evaluations, university admissions, and standardized tests such as TOEFL and IELTS \citep{zupanc2018increasing}. Evaluation is typically conducted by educators using structured scoring rubrics designed to ensure consistency and objectivity. Scoring rubrics generally follow either a holistic or analytic approach. The holistic scoring assigns a single overall score based on general writing quality, while the analytical scoring evaluates multiple dimensions, such as organization, grammar, vocabulary, and cohesion, to provide more detailed feedback on specific writing traits \citep{li2024automated}.

Manual evaluation presents several challenges, especially when large volumes of essays must be evaluated within limited timeframes. Even with clear rubrics, human scoring remains vulnerable to inconsistencies, subjectivity, and rater bias, which may compromise reliability and fairness \citep{uto2020robust}. To address these limitations, Automated Essay Scoring (AES) systems have been developed as a practical and scalable solution. By leveraging techniques from Natural Language Processing (NLP) and Machine Learning (ML), AES systems automatically evaluate written texts with minimal human intervention \citep{uto2021review, misgna2025survey}. This automation enables consistent and objective scoring, reduces evaluation time and cost,  and supports educational feedback by enabling rapid, large-scale assessments \citep{misgna2025survey, ramesh2022automated}.

Early AES systems relied heavily on handcrafted features extracted from essays, such as lexical, syntactic, grammatical, and content-specific indicators. Although effective to some extent, these feature engineering approaches require significant manual effort and lack flexibility when adapting to new traits or tasks \citep{misgna2025survey}. To address these limitations, neural network-based methods were introduced, enabling automatic learning of feature representations directly from text data through word embeddings. These embeddings encode semantic similarity based on context, eliminating the need for handcrafted features \citep{misgna2025survey}.

Despite their success, deep learning models like Convolutional Neural Networks (CNNs) and Recurrent Neural Networks (RNNs) often fail to capture global dependencies in a corpus, which are essential for holistic understanding. They mostly operate on local sequences of words and lack mechanisms to model relationships across multiple documents \citep{yao2019graph}.

To address this, Graph Neural Networks (GNNs) have gained attention. GNNs can model long-range dependencies and global relationships by representing texts as graphs \citep{cai2018comprehensive}. One popular architecture, Graph Convolutional Networks (GCNs), introduced by Kipf and Welling \citep{kipf2017semi}, has shown promising results in NLP tasks including text classification \citep{yao2019graph, liu2020tensor}.

Recent studies have applied GCNs to AES, demonstrating their effectiveness in capturing relationships between essays and enhancing feature extraction. For instance, Ait et al. \citep{ait2020graph} used TF-IDF and Word2Vec to represent nodes, capturing both term importance and semantic meaning. Tan et al. \citep{tan2023automatic} employed one-hot encoding to create binary vector representations, while Ma et al. \citep{ma2021enhanced} used GloVe embeddings to represent documents, sentences, and tokens, enabling more comprehensive modeling of semantic structures. These approaches differ primarily in how graph nodes are represented through word embeddings. Although AES models have shown promising results, several challenges and limitations remain. The following outlines these key issues.

% Although AES models based on GCNs have achieved promising results, several challenges and limitations remain.

\begin{itemize}
    \item Traditional word embedding methods, such as TF-IDF, Word2Vec, and GloVe, produce static representations that fail to capture contextual meaning and struggle with polysemy. Consequently, the same embedding is assigned to words used in different contexts, limiting their effectiveness in tasks like AES. Pre-trained Transformer models address these shortcomings by generating dynamic, context-aware embeddings that capture rich semantic and syntactic information \citep{she2022joint}. However, while Transformers excel at modeling intra-textual context, they are less effective at capturing relationships across multiple texts—such as similarities and dependencies among essays. To address this limitation, recent approaches have combined Transformer-based models with GCNs, an integration that has proven effective across various NLP tasks, including text classification \citep{she2022joint, gao2021gating, lu2020vgcn, albadani2022transformer}. This hybrid approach leverages the strengths of both architectures: the contextual depth of Transformers and the relational modeling capabilities of GCNs, resulting in a more robust and comprehensive solution for AES and other NLP tasks.
    \item Despite the advantages of GCNs, a notable limitation remains: the equal treatment of all neighboring nodes, which may lead to suboptimal modeling of node relevance. To overcome this issue, Graph Attention Networks (GATs), proposed by Veličković et al. \citep{velivckovic2017graph}, incorporate an attention mechanism that assigns learnable weights to neighboring nodes based on their importance. GATs have achieved promising results across various NLP tasks, such as text classification and sentiment analysis, demonstrating their effectiveness in capturing complex relationships within textual data \citep{haitao2022text, vrahatis2024graph}. By weighting the importance of neighboring nodes, GATs enable models to focus more effectively on the most relevant connections—an advantage particularly beneficial for AES, where the strength of relationships between essays can vary.
    \item Most existing AES systems rely on holistic scoring, which assigns a single score to represent the overall quality of an essay. Although this approach simplifies the scoring process, it fails to assess specific aspects of writing such as grammar, vocabulary, coherence, and organization. Consequently, it limits the ability to provide detailed and targeted feedback—an essential component in educational contexts, especially for second-language learners. Holistic scoring does not reflect the full range of a student's writing abilities, reducing its effectiveness in identifying strengths and weaknesses. Despite these drawbacks, much of the prior research has focused on holistic evaluation, often overlooking the advantages of analytic scoring, which assigns multiple scores across distinct writing dimensions and offers more informative feedback to support writing improvement \citep{li2024icle++}.
    \item The Automated Student Assessment Prize (ASAP) dataset, released as part of a 2012 Kaggle competition, has become the primary benchmark for evaluating AES models, but it has notable limitations that constrain its usefulness for broader AES research. First, the dataset lacks diversity, consisting of essays written by U.S. students in grades 7 to 10, most of whom are native English speakers. This raises concerns about generalizability, particularly when applying AES models to essays written by second-language learners or students from different educational backgrounds. Second, the structure of ASAP limits its suitability for evaluating models in cross-prompt scoring. AES researchers have traditionally relied on within-prompt scoring, where models are trained and tested on the same prompt. However, this setup is often impractical in real-world applications, as AES models tend to perform poorly on unseen prompts unless retrained on new data. In response, researchers have begun exploring cross-prompt scoring, a more challenging setting where models are trained on multiple prompts and tested on essays from prompts not seen during training. Although ASAP contains eight prompts across different genres (narrative, persuasive, and source-based), using it for cross-prompt evaluation introduces complications, as scoring rubrics and writing expectations vary by genre \citep{li2024icle++, li2024automated}. Recently, a new dataset has been introduced to address the limitations of the ASAP dataset. Composed of essays written by English Language Learners, this dataset features a larger number of prompts within the same genre \citep{crossley2023english}. These characteristics make it more suitable for training and evaluating robust AES models. Further details about the dataset are provided in the Experiments section.
\end{itemize}

In conclusion, although current AES models have made considerable advancements, there is still potential for further development in several areas. By tackling these challenges, it is possible to develop more efficient and adaptable AES models that offer deeper, more comprehensive feedback and better meet the varied requirements of educational environments.

To address these shortcomings, this paper proposes the TransGAT method, which effectively combines the contextual representation power of pre-trained Transformers with the relational modeling capabilities of GATs. It also leverages an enhanced dataset to train a more robust AES model capable of assessing essays across multiple dimensions. Additionally, this study addresses another research gap by combining cross-prompt essay scoring with multi-dimensional essay scoring. This study is the first to integrate Transformer-based embeddings with GATs for AES while jointly addressing cross-prompt and multi-dimensional scoring.

This paper is organized as follows: Section~\ref{sec:related_work} provides a review of related work in the field of AES. Section~\ref{sec:methodology} presents the proposed methodology, detailing the integration of Transformer models and GAT. Section~\ref{sec:experiments} summarizes the experiments, including dataset details, evaluation metric used to assess performance, experimental setup, results obtained, and comparison to baseline studies. Finally, Section~\ref{sec:conclusion} discusses the conclusions and suggests directions for future research.

\section{Related Work}
\label{sec:related_work}
This section categorizes previous studies on the AES task into two groups based on the rating scales used for writing assessment: holistic scoring and analytic scoring.

\subsection{Holistic Scoring}
\label{subsec:holistic_studies}
Research on holistic essay scoring has progressed from traditional machine learning methods to deep neural networks and, more recently, to Graph Convolutional Networks (GCNs). Traditional machine learning methods relied on handcrafted features such as vocabulary, sentence length, grammatical errors, and syntactic complexity \citep{madala2018empirical, sharma2020automated}. Advanced features, including parse tree depth and type-token ratio, have also been explored \citep{salim2019automated, janda2019syntactic, doewes2020structural}. 

Deep learning methods introduced word embeddings to automate feature extraction, initially using static embeddings like GloVe and Word2Vec with CNNs, GRUs, LSTMs, and Bi-LSTMs \citep{chen2020automatic, li2018coherence, cai2019automatic, zhu2020automated, xia2019automatic, muangkammuen2020multi, chen2018relevance}. Attention mechanisms further enhanced performance \citep{chen2018relevance}. Transformer-based models, such as BERT, RoBERTa, and DeBERTa, provided contextualized embeddings that outperformed static embeddings in AES tasks and yielded improved results when combined with LSTMs, CNNs, Capsule Networks, and Bi-LSTMs \citep{wangkriangkri2020comparative, wang2023study, sharma2021feature, yang2021automated, yang2020enhancing, mayfield2020should, susanto2023development, beseiso2021novel}. Hybrid approaches integrated both static and contextual embeddings to enhance performance \citep{li2023automatic, beseiso2020empirical, zhou2021self}. However, most deep learning models primarily focus on local word sequences without explicitly capturing global word relationships \citep{yao2019graph}.

GCNs have emerged as a promising method for capturing global relationships within textual data \citep{ren2022graph, bhatti2023deep}. Early studies applied GCNs to short-answer scoring, leveraging one-hot encoding, TF-IDF, and Word2Vec embeddings \citep{tan2023automatic, ait2020graph}. Later research extended GCNs to essay scoring using GloVe embeddings \citep{ma2021enhanced}. However, most prior work has primarily focused on holistic scoring, evaluating overall essay quality without addressing specific writing aspects.

\subsection{Analytic Scoring}
\label{subsec:analytic_studies}
Initial research in AES primarily utilized traditional machine learning algorithms combined with manually engineered features. In recent years, however, numerous studies have shown that neural network methods can more effectively capture the deep semantics of essays. For example, Li et al. \citep{li2024conundrums} trained a simple neural network using extracted features as input and achieved state-of-the-art performance. Mathias et al. \citep{mathias2020can} compared a feature-engineering system, a string kernel-based model, and an attention-based neural network using GloVe pre-trained word embeddings to automatically extract features for scoring various essay traits, finding that the neural network delivered the best results. The introduction of deep learning further automated feature extraction through word embeddings, with models such as Convolutional Neural Networks (CNNs), Recurrent Neural Networks (RNNs), Gated Recurrent Units (GRUs), Long Short-Term Memory networks (LSTMs), and Bidirectional LSTMs (BiLSTMs) showing promising results \citep{hussein2020trait, ridley2021automated, shin2022evaluating, do2023prompt, chen2023pmaes}.

More recently, many AES studies have utilized transformer-based pre-trained language models—such as BERT, DistilBERT, RoBERTa, and DeBERTa—to obtain essay embeddings, followed by model fine-tuning to optimize performance \citep{lee2021automated, xue2021hierarchical, ormerod2022mapping, lohmann2024neural, sun2024automatic, chen2024multi}. Several studies have enhanced these transformer models by integrating them with additional deep-learning architectures. For example, Lee et al. \citep{cho2024dual} incorporated a CNN with the output from BERT to improve local feature extraction, while another study \citep{lee2023nc2t} combined RoBERTa with an enhanced Bidirectional GRU (BiGRU) to boost scoring accuracy. Recent work has also begun to explore the use of large language models (LLMs), which have shown strong performance in AES tasks \citep{xiao2025human}.

However, most of these approaches primarily focus on local word sequences and often overlook the global relationships between words \citep{yao2019graph}. To address this limitation, GCNs have emerged as a promising technique for capturing global dependencies within textual data \citep{ren2022graph, bhatti2023deep}. Building on this, the present study integrates Transformer-based contextual embeddings with GCNs for multidimensional essay scoring. Transformer models, such as BERT and DeBERTa, provide rich, context-dependent representations of essays that effectively capture semantic features. By feeding these embeddings into a GCN, the proposed hybrid approach leverages the strengths of both architectures: the deep semantic understanding from Transformers and the global relational modeling from GCNs. This combination enhances predictive performance and contributes to more accurate and robust AES outcomes.

\section{Methodology}
\label{sec:methodology}
\begin{figure}[H]
    \centering
    \fbox{%
      \includegraphics[width=0.95\textwidth]{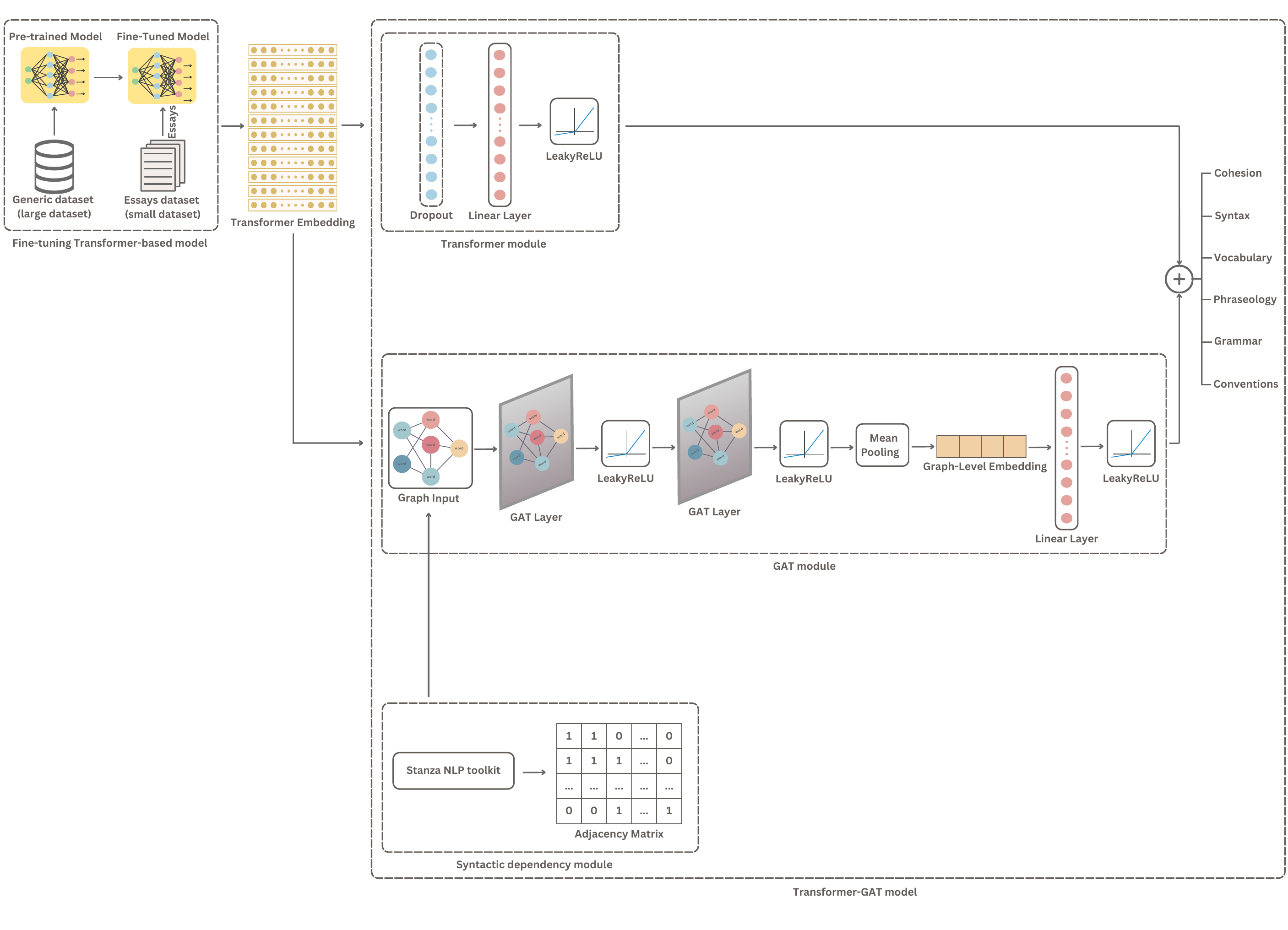}%
    }
    \caption{The overall architecture of the TransGAT method.}
    \label{proposedMethod}
\end{figure}

This section presents TransGAT, a hybrid method developed for analytic AES. The method consists of two main components: a fine-tuned Transformer-based model and a Transformer–GAT model. The first component extracts contextual embeddings that capture the semantic features of each essay. The second component, the Transformer–GAT model, extends the fine-tuned Transformer by integrating it with a GAT to effectively model both semantic content and syntactic structure. Within this component, two-stream predictions are performed by pairing each Transformer with a separate GAT. In each pair, the first stream generates essay-level predictions using the Transformer, while the second uses Transformer token representations within GAT, where edges are constructed based on syntactic dependencies. A detailed explanation of each component is provided in the following subsections, with the overall architecture illustrated in Figure \ref{proposedMethod}.

\subsection{Fine-tuning Transformer-based model}
\label{subsec:fine_tuning}
Transformer-based language models like BERT have demonstrated remarkable performance in NLP tasks \citep{gillioz2020overview}, building upon the Transformer architecture originally proposed by Vaswani et al. \citep{vaswani2017attention}. Unlike traditional sequential models, the Transformer uses an encoder-decoder structure that processes entire input sequences simultaneously. Although it does not inherently encode word order, it incorporates positional encoding to retain sequence information, ensuring that word representations vary based on their positions. The self-attention mechanism allows the model to capture dependencies between all tokens in a sequence \citep{vaswani2017attention}.

These models are highly effective at learning general-purpose language representations from large-scale unlabeled corpora, which can be transferred to various downstream tasks through transfer learning. This process involves unsupervised pre-training on extensive text data, followed by supervised fine-tuning on task-specific datasets to adapt the model to the target domain. This strategy often yields superior performance compared to training models from scratch \citep{schomacker2021language}.

This study evaluates the performance of three Transformer-based pre-trained models on the AES task, using the large version of each model for comparison. Each model was fine-tuned on the AES datasets to generate embeddings tailored to the data, which then served as the foundation for further processing.

\begin{enumerate}
\item BERT, short for "Bidirectional Encoder Representations from Transformers," is a foundational model in natural language processing that introduced a deep bidirectional approach to language understanding. The large version of BERT, known as BERT-large, consists of 24 layers and 340 million parameters. It is pre-trained on large corpora using masked language modeling and next-sentence prediction tasks, allowing it to capture context from both directions in a sentence. BERT-large has achieved strong performance across a wide range of NLP benchmarks \citep{devlin2019bert}. 

\item RoBERTa, short for "Robustly Optimized BERT Pre-training Approach," replicates the original BERT by optimizing hyperparameters and significantly increasing the size of the training data. The RoBERTa-large model, with 24 layers, is pre-trained on five diverse English-language datasets totaling over 160GB of text. RoBERTa has achieved state-of-the-art results in several downstream tasks across multiple benchmarks \citep{liu2019roberta}.

\item DeBERTa, short for "Decoding-enhanced BERT with Disentangled Attention," further refines the BERT and RoBERTa models by introducing two novel techniques: a disentangled attention mechanism and an enhanced masked decoder. The large version of DeBERTa, DeBERTa-large, improves pretraining efficiency and model performance \citep{he2021debertav3}.
\end{enumerate}

\subsection{Transformer-GAT model}
\label{subsec:transformer_gat}
The model consists of four interconnected modules: a Transformer module, a syntactic dependency module, a GAT module, and a fusion module. The Transformer module leverages a fine-tuned Transformer model (described earlier) to extract contextualized embeddings from the input text, which are then passed through a dense layer to generate essay-level predictions. At the same time, the syntactic dependency module employs the Stanza toolkit to parse the syntactic structure of the text and to represent it as an adjacency matrix. In the GAT module, the contextualized word embeddings obtained from the fine-tuned Transformer model are used as node features, while the edges of the graph are defined by syntactic dependencies extracted by the syntactic dependency module. This graph is processed through two GAT layers, followed by a global mean pooling to generate a graph-level representation for prediction. Finally, the outputs of the Transformer and the GAT modules are fused to produce the final analytic scores.

\subsubsection{Transformer module}
\label{subsubsec:transformer_module}
This module uses the fine-tuned transformer model from the first component, which has already been trained on the essay dataset. At this stage, the Transformer model is frozen, which means that its weights are not updated during this step. Freezing the model allows us to utilize the pre-learned representations obtained during fine-tuning without further modification.

The Transformer module extracts contextualized embeddings for the input text. Specifically, it uses the embedding of the special \texttt{[CLS]} token, which serves as a summary representation of the entire essay. This \texttt{[CLS]} token embedding, denoted as \( h_{\text{[CLS]}} \in \mathbb{R}^d \), is then passed through a dense (fully connected) layer followed by a non-linear activation function, LeakyReLU, to generate predictions at the essay level for various analytic aspects.

Formally, the dense layer applies a linear transformation using a weight matrix \( W \in \mathbb{R}^{k \times d} \) and a bias vector \( b \in \mathbb{R}^k \), where \(d\) is the embedding dimension and \(k\) is the number of analytic aspects to predict. The output vector \( y \in \mathbb{R}^k \) is computed as:

\begin{equation}
y = \text{LeakyReLU}(W h_{\text{[CLS]}} + b).
\end{equation}

% Here, the output vector \( y \) contains \( k \) values, each representing the predicted score for a specific analytic aspect of the essay. For example, if the analytic aspects include grammar, cohesion, vocabulary, syntax, phraseology, and conventions, then \( k = 6 \), and each element of \( y \) corresponds to the model’s prediction for one of these aspects.

Each element of the output vector \( y \) corresponds to a predicted score for one of the \( k \) analytic aspects of the essay (such as grammar, cohesion, vocabulary, etc.). This approach enables the model to take advantage of the rich semantic information captured by the Transformer’s \texttt{[CLS]} token embedding while producing multi-dimensional outputs aligned with the analytical scoring criteria.

\subsubsection{Syntactic dependency module}
\label{subsubsec:syntactic_module}
This module converts each essay into a syntactic text graph to incorporate structural linguistic information. The Stanza NLP toolkit is employed to analyze the syntactic dependency tree of the input text, which identifies grammatical relationships between words such as subject--verb or modifier--noun connections. This tree is then transformed into an adjacency matrix, where each entry indicates the presence of a direct syntactic relationship between a pair of words. 

Based on this matrix, a graph \( G = (V, E) \) is constructed, where nodes \( V \) (with \( |V| = n \)) correspond to the tokens of an essay and $x \in \mathbb{R}^{n \times d}$ denote the word embeddings of the input tokens generated by the fine-tuned Transformer model, and edges \( E \) represent syntactic dependencies between the tokens. A dependency adjacency matrix $A \in \mathbb{R}^{n \times n}$ is built, where $A_{ij} = 1$ if there is a syntactic relation between tokens $i$ and $j$, and 0 otherwise. The adjacency matrix is then converted into a sparse coordinate format for graph construction.

\begin{equation}
A_{ij} =
\begin{cases}
1 & \text{if token } i \text{ is syntactically connected to token } j, \\
0 & \text{otherwise}.
\end{cases}
\end{equation}

The entire graph construction process is illustrated in Algorithm 1.
\begin{algorithm}[htp!]
\caption{Graph Construction Using Dependency Relations}
\begin{algorithmic}[1]
 \renewcommand{\algorithmicrequire}{\textbf{Input:}}
 \renewcommand{\algorithmicensure}{\textbf{Output:}}
 \REQUIRE Set of tokens \( T = \{t_1, t_2, \dots, t_n\} \), with their corresponding dependency relations
 \ENSURE Graph \( G = (V, E) \), where \( V \) is the set of tokens and \( E \) is the set of edges based on syntactic relations
 \\ \textit{Initialization:}
 \STATE Construct a dependency adjacency matrix \( A \in \mathbb{R}^{n \times n} \) such that:
 \[
 A_{ij} =
 \begin{cases}
 1 & \text{if token } i \text{ is syntactically connected to token } j, \\
 0 & \text{otherwise}
 \end{cases}
 \]
 \\ \textit{Graph Construction:}
 \FOR{each pair of tokens \( (i, j) \)}
    \IF{\( A_{ij} = 1 \)}
        \STATE Add edge \( (t_i, t_j) \) to the graph \( G \)
    \ENDIF
 \ENDFOR
 \\
 \RETURN \( G \)
\end{algorithmic}
\end{algorithm}

To enrich the graph structure, the syntactic graph is treated as undirected. This representation enables the model to capture syntactic structure more effectively, which is crucial for assessing analytic aspects such as grammar, clarity, and sentence complexity.

\subsubsection{GAT module}
\label{subsubsec:gat_module}
After constructing the text graph for each document, information is propagated and updated between nodes using the GAT to obtain node representations that contain both text syntax and sequence information. Unlike the standard self-attention mechanism, which assigns attention weights globally across all nodes, the graph attention mechanism operates locally and does not require knowledge of the entire graph structure. This allows the model to flexibly assign different weights to neighboring nodes and to parallelize computation across all nodes efficiently.

The GAT accepts two inputs: the node features, which are the word embeddings produced by the fine-tuned Transformer module,
\[
H = \{ \mathbf{h}_1, \mathbf{h}_2, \ldots, \mathbf{h}_n \},
\]
where each \( \mathbf{h}_i \in \mathbb{R}^m \), \( n \) is the number of nodes, and \( m \) is the number of features per node. The second input is the adjacency matrix \( A \), obtained through the syntactic dependency module, which captures neighbor relationships among nodes.

Specifically, for a node \( i \) and one of its neighbors \( j \), the attention coefficient \( e_{ij} \) is computed as:
\begin{equation}
    e_{ij} = \text{LeakyReLU}\left( a^\top [W \mathbf{h}_i \, \| \, W \mathbf{h}_j] \right),
\end{equation}
where:
\begin{itemize}
    \item \( W \) is a learnable weight matrix,
    \item \( a \) is a learnable attention vector,
    \item \( ^\top \) denotes transposition,
    \item \( \| \) indicates vector concatenation,
    \item \texttt{LeakyReLU} is the non-linear activation function.
\end{itemize}

The attention coefficients are then normalized using the softmax function:
\begin{equation}
    \alpha_{ij} = \frac{\exp(e_{ij})}{\sum_{k \in \mathcal{N}(i)} \exp(e_{ik})},
\end{equation}
where \( \mathcal{N}(i) \) is the set of neighbors of node \( i \).

Using the normalized attention weights, the updated representation of node \( i \) is computed as:
\begin{equation}
    \mathbf{h}'_i = \sigma\left( \sum_{j \in \mathcal{N}(i)} \alpha_{ij} W \mathbf{h}_j \right),
\end{equation}
where \( \sigma \) denotes the LeakyReLU activation function.

The GAT layer produces updated node representations:
\[
H' = \{ \mathbf{h}'_1, \mathbf{h}'_2, \ldots, \mathbf{h}'_n \},
\]
where each \( \mathbf{h}'_i \in \mathbb{R}^m \).

After two successive GAT layers, the updated node features are aggregated into a fixed-size graph-level representation using global mean pooling:
\begin{equation}
    \mathbf{h}_G = \text{MeanPooling}\left( \{ \mathbf{h}''_i \}_{i=1}^{n} \right),
\end{equation}
where \( \mathbf{h}''_i \) denotes the node representation after the second GAT layer, and \( n \) is the number of nodes.

This operation computes the average of all node embeddings across each feature dimension, resulting in a single graph-level vector representing the entire essay. The graph-level representation is then passed through a fully connected linear layer followed by a LeakyReLU activation function to produce the final prediction:
\begin{equation}
    \mathbf{y} = \sigma(W_2 \mathbf{h}_G + b_2),
\end{equation}
where \( W_2 \) and \( b_2 \) are trainable parameters, \( \mathbf{h}_G \) is the graph-level embedding, and \( \sigma \) denotes the LeakyReLU activation function.

\subsubsection{Fusion module}
\label{subsubsec:fusion_module}
The final essay scores are computed as a combination of both the Transformer and GAT modules:
\begin{equation}
    \hat{y} = s_1 + s_2,
\end{equation}
where \( \hat{y} \in \mathbb{R}^{C} \) represents the predicted scores across \( C \) assessment criteria.

By incorporating Transformer-based contextual embeddings alongside d-\newline ependency-parsing-based graph attention, the model effectively captures both semantic meaning and syntactic structure, resulting in a more robust AES system.

\subsection{Training Procedure}
\label{subsec:training_procedure}
Both the Transformer and Transformer-GAT models are trained using Mean Squared Error (MSE) loss:
\begin{equation}
    \mathcal{L}_{\text{MSE}} = \frac{1}{6} \sum_{i=1}^6 (y_i - \hat{y}_i)^2.
\end{equation}

where:
\begin{itemize}
    \item $y_i$ is the ground-truth score for the $i$-th analytic trait,
    \item $\hat{y}_i$ is the predicted score for the $i$-th analytic trait,
    \item $i$ indexes the six traits: cohesion, syntax, vocabulary, phraseology, grammar, and conventions,
    \item $\mathcal{L}_{\text{MSE}}$ is the average loss across all six traits.
\end{itemize}

\section{Experiments and Discussion}
\label{sec:experiments}
This section starts by introducing the datasets utilized in the study, then outlines the approach used for performance evaluation. It proceeds with a description of the experimental setup. Following that, the experimental results are presented along with an in-depth analysis and interpretation. Lastly, the discussion compares the performance of the proposed TransGAT method against baseline models.

\subsection{Dataset}
\label{subsec:dataset}
The English Language Learner Insight, Proficiency, and Skills Evaluation (ELLIPSE) corpus~ \citep{crossley2023english} is a publicly available dataset containing approximately 6,500 essays written by English Language Learners (ELLs) in grades 8–12. These essays were collected from statewide standardized assessments in the United States and span 29 distinct writing prompts. Each essay is annotated with both holistic and analytic proficiency scores, including evaluations of cohesion, syntax, vocabulary, phraseology, grammar, and conventions. On average, the essays contain around 430 words, most ranging between 250 and 500. Scoring is performed on a scale from 1.0 to 5.0, in 0.5-point increments. Two trained human raters evaluated each essay to ensure scoring reliability. This corpus addresses the limitations of previous analytic datasets, such as ASAP++ \citep{mathias2018asap++}, which extended the original ASAP dataset to support analytic scoring but retained many of its constraints.

\subsection{Performance Evaluation}
\label{subsec:qwk}
Quadratic Weighted Kappa (QWK) is a statistical metric used to assess the level of agreement between predicted and actual scores, making it especially well-suited for evaluating AES systems. Unlike basic accuracy measures, QWK accounts for the extent of disagreement between predicted and true scores, offering a more detailed evaluation of a model’s performance \citep{ramnarain2022similarity}.

The QWK score is computed using the following formula \citep{ramnarain2022similarity}:

\begin{equation}
QWK = 1 - \frac{\sum_{i,j} w_{i,j} O_{i,j}}{\sum_{i,j} w_{i,j} E_{i,j}},
\end{equation}
where:
\begin{itemize}
    \item \( O_{i,j} \) denotes the observed agreement matrix, which reflects how often each predicted score aligns with the actual score.
    \item \( E_{i,j} \) refers to the expected agreement matrix, indicating the level of agreement that would be expected by random chance.
    \item \( w_{i,j} \) represents the weight matrix, usually defined as:
    \begin{equation}
    w_{i,j} = \frac{(i - j)^{2}}{(N - 1)^{2}},
    \end{equation}
    where \( i \) and \( j \) correspond to the predicted and actual score categories, and \( N \) is the total number of distinct score levels.
\end{itemize}

QWK scores range from -1 to 1. A score of 1 reflects perfect agreement between the predicted and actual values, while a score of 0 suggests agreement no better than random chance. Negative values imply systematic disagreement. QWK incorporates a quadratic penalty, which means that larger differences between predicted and actual scores are penalized more heavily. This feature is especially important in educational settings like AES, where accurate scoring is essential, and significant errors can undermine the model’s credibility. As outlined in Table~\ref{tab:Kappa_values}, QWK scores can be interpreted to assess the strength of agreement, with values below 0 indicating no agreement and those between 0.81 and 1.00 suggesting near-perfect agreement \citep{ramnarain2022similarity}.

\begin{table}[htp!]
\centering
\caption{Interpretation of Kappa Values}
\begin{tabular}{|c|c|}
\hline
\textbf{Kappa Value} & \textbf{Interpretation} \\ \hline
$<$ 0 & No agreement \\ \hline
0.01 - 0.20 & Slight agreement \\ \hline
0.21 - 0.40 & Fair agreement \\ \hline
0.41 - 0.60 & Moderate agreement \\ \hline
0.61 - 0.80 & Substantial agreement \\ \hline
0.81 - 1.00 & Almost perfect agreement \\ \hline
\end{tabular}
\label{tab:Kappa_values}
\end{table}

\subsection{Experimental Setup}
\label{subsec:setup}
The experimental setup includes configurations for both Transformer-based models and GAT models. The maximum input sequence length was determined based on model capacity: $1024$ tokens for DeBERTaV3 to leverage its ability to process longer inputs, and $512$ tokens for other models, in line with their architectural constraints. The tokenizers were extended to recognize special tokens representing paragraph breaks (\textbackslash n) and double spaces, which are important for preserving the linguistic structure of essays. Training and evaluation were conducted using a batch size of $4$, and the models that achieved the best performance on the validation set were saved for subsequent analysis.

The GAT model comprised two layers and four attention heads and employed the LeakyReLU activation function. Both model types were optimized using the AdamW optimizer. The learning rate was set to $1e{-5}$ for the Transformer models and $1e{-3}$ for the Transformer-GAT model. A weight decay of $1e{-2}$ was applied only to the Transformer models. Both setups utilized a cosine learning rate scheduler and were trained for six epochs. No dropout was applied during training, as this was found to improve regression performance.

\subsection{Experimental Results}
\label{subsec:result}
Extensive experiments were conducted by systematically varying several key factors to optimize model performance. These included using different dependency parsing tools—such as spaCy and Stanza—to construct the graph structure, tuning dropout rates to prevent overfitting, applying distinct learning rates for the Transformer and GAT components to balance their training, experimenting with various optimizers (e.g., Adam, AdamW), and testing different learning rate schedulers. Additionally, the number of GAT layers and attention heads, choice of activation functions (e.g., ReLU, LeakyReLU), and batch sizes were adjusted. These variations helped identify the optimal configuration for each analytic scoring dimension. The best results from these experiments are presented in Table~\ref{tab:analytic-results}, which reports QWK scores on the ELLIPSE dataset across six analytic scoring dimensions: Cohesion, Syntax, Vocabulary, Phraseology, Grammar, and Conventions.

\begin{table}[htp!]
\centering
\caption{QWK on ELLIPSE, with best scores in bold.}
\label{tab:analytic-results}
\resizebox{\columnwidth}{!}{%
\begin{tabular}{|c|c|c|c|c|c|c|c|}
\hline
\textbf{Model} & \textbf{Avg. QWK} & \textbf{Cohesion} & \textbf{Syntax} & \textbf{Vocabulary} & \textbf{Phraseology} & \textbf{Grammar} & \textbf{Conventions} \\
\hline
BERT-large-GAT & 0.808 & 0.799 & 0.809 & 0.796 & 0.782 & 0.861 & 0.801 \\
RoBERTa-large-GAT & \textbf{0.854} & 0.852 & 0.852 & \textbf{0.825} & \textbf{0.861} & \textbf{0.877} & \textbf{0.859} \\
DeBERTaV3-large-GAT & 0.833 & \textbf{0.877} & \textbf{0.853} & 0.813 & 0.836 & 0.872 & 0.744 \\
\hline
\end{tabular}
}
\end{table}

RoBERTa-large-GAT achieved the highest average QWK score of $0.854$, reflecting its top performance in four out of six analytic dimensions among the GAT variants: Vocabulary ($0.825$), Phraseology ($0.861$), Grammar ($0.877$), and Conventions ($0.859$). In contrast, the DeBERTaV3-large-GAT model demonstrated strong performance in the remaining two dimensions, Cohesion ($0.877$) and Syntax ($0.853$), but its lower score in Conventions ($0.744$) contributed to a reduced overall average.

These findings underscore RoBERTa's ability to capture complex linguistic features, making it more suitable for this task than DeBERTaV3. One possible explanation is that RoBERTa was pre-trained on a significantly larger corpus (160GB of text) compared to DeBERTaV3's 78GB, which may enhance its effectiveness on AES tasks—despite DeBERTaV3 generally outperforming RoBERTa in other benchmarks \citep{he2020deberta}.

\subsection{Discussion}
\label{subsec:discussion}
Table~\ref{tab:ellipse-comparison} presents a comparative analysis of the proposed TransGAT method against two baseline studies, as these are the only studies that utilized the ELLIPSE dataset.

\begin{table}[htp!]
\centering
\caption{Comparison of baselines vs. TransGAT, with best scores in bold.}
\label{tab:ellipse-comparison}
\resizebox{\textwidth}{!}{%
\begin{tabular}{|c|c|c|c|c|c|c|c|c|c|}
\hline
\textbf{Model Group} & \textbf{Study} & \textbf{Model} & \textbf{Avg. QWK} & \textbf{Cohesion} & \textbf{Syntax} & \textbf{Vocabulary} & \textbf{Phraseology} & \textbf{Grammar} & \textbf{Conventions} \\
\hline\hline
\multirow{2}{*}{Transformers Models} 
& Sun et al.~\cite{sun2024automatic} & roberta-base & 0.825 & 0.81 & 0.83 & \textbf{0.84} & 0.80 & 0.83 & 0.84 \\
\cline{2-10}
& Chen et al.~\cite{chen2024multi} & debertaV3-base & 0.685 & 0.63 & 0.69 & 0.67 & 0.69 & 0.72 & 0.71 \\
\hline\hline
\multirow{3}{*}{Transformer-GAT}  
& \multirow{3}{*}{TransGAT method} & bert-large-GAT & 0.808 & 0.799 & 0.809 & 0.796 & 0.782 & 0.861 & 0.801 \\
\cline{3-10}
& & roberta-large-GAT & \textbf{0.854} & 0.852 & 0.852 & 0.825 & \textbf{0.861} & \textbf{0.877} & \textbf{0.859} \\
\cline{3-10}
& & debertaV-large-GAT & 0.833 & \textbf{0.877} & \textbf{0.853} & 0.813 & 0.836 & 0.872 & 0.744 \\
\hline
\end{tabular}%
}
\end{table}

The baseline studies, represented by Sun et al. \citep{sun2024automatic} and Chen et al. \citep{chen2024multi}, employ RoBERTa-base and DeBERTaV3-base models, respectively. Sun et al.'s RoBERTa-base achieve relatively strong performance across most dimensions, with a notably high score in Vocabulary ($0.84$). Conversely, Chen et al.'s DeBERTaV3-base, despite being more recent, performs relatively lower across all dimensions, indicating a less effective adaptation to this task.

The TransGAT method, which integrates GAT with Transformer-based models, consistently improves performance over the baseline models. Specifically, the RoBERTa-large-GAT variant shows the strongest results among the baseline studies, achieving the highest scores in three dimensions: Phraseology ($0.861$), Grammar ($0.877$), and Conventions ($0.859$). Meanwhile, the DeBERTaV3-large-GAT variant excels in Cohesion ($0.877$) and Syntax ($0.853$), surpassing the RoBERTa-large-GAT model in these dimensions.

Although the TransGAT method generally outperforms baseline models across most analytic dimensions, it shows relatively lower performance in the Vocabulary dimension compared to Sun et al.'s RoBERTa-base model ($0.84$). This could be attributed to the fact that vocabulary scoring often relies heavily on surface-level lexical richness and diversity, which the baseline RoBERTa-base model may capture more directly due to its simpler architecture and focus on token-level representations. In contrast, the integration of GAT in TransGAT emphasizes relational and contextual information across essays, which may reduce attention to details lexical features critical for vocabulary scoring.

Overall, the results highlight the advantage of incorporating GAT with Transformer architectures. The TransGAT method effectively captures richer contextual and relational information between essays, enabling superior modeling of complex linguistic features critical to AES. This is particularly evident in the RoBERTa-large-GAT model’s performance, which outperforms the baseline RoBERTa-base model across almost all analytic dimensions.

\section{Conclusion}
\label{sec:conclusion}

This study introduced TransGAT, a novel method for analytic AES that effectively combines the strengths of Transformer-based models and GAT. The proposed approach consists of two key components: a fine-tuned Transformer that generates rich contextual embeddings, and a GAT that incorporates syntactic structure through attention-based graph modeling. This hybrid design allows TransGAT to capture both semantic content and intra-essay structural dependencies, enabling a more accurate and nuanced evaluation of writing quality.

A key innovation in TransGAT is its two-stream prediction mechanism, where one stream produces essay-level predictions using the Transformer, while the other leverages token-level Transformer outputs within a GAT framework, using syntactic dependencies to guide edge construction. This dual approach allows the model to focus dynamically on the most relevant linguistic and structural features, which is particularly important in AES, where variations in organization, grammar, and coherence affect scoring.

Extensive experiments conducted on analytic AES tasks demonstrated that TransGAT consistently outperforms baseline models across multiple scoring dimensions. The attention-based relational modeling within GAT enhances the model’s ability to differentiate between essays with subtle variations in quality, leading to more reliable predictions. These results underscore the benefit of integrating deep language representations with graph-based relational reasoning in educational NLP.

Looking forward, future work will explore expanding TransGAT with heterogeneous graph structures that include both word-level and essay-level nodes, allowing the model to capture more complex interactions across and within essays. Another promising direction is adapting TransGAT for morphologically rich languages such as Arabic. The unique linguistic features of Arabic—including its root-based morphology, optional diacritics, and diverse dialects—pose challenges for conventional NLP pipelines. Addressing these through Arabic-specific tokenizers, morphological analyzers, and syntactic parsers will help evaluate the cross-linguistic generalizability of the proposed approach and contribute to the development of AES tools for underrepresented languages.

\section*{Acknowledgments}
The project was funded by KAU Endowment (WAQF) at king Abdulaziz University, Jeddah, Saudi Arabia. The authors, therefore, acknowledge with thanks WAQF and the Deanship of Scientific Research (DSR) for technical and financial support.
 \bibliographystyle{elsarticle-harv} 
 \bibliography{TransGAT}




% %% For authoryear reference style

% \bibitem[Lamport(1994)]{lamport94}
%   \textit{\LaTeX: a document preparation system},
%   Addison Wesley, Massachusetts,

% \end{thebibliography}
\end{document}